\documentclass[a4paper]{svproc}

\usepackage{type1cm}        
\usepackage{makeidx}         
\usepackage{graphicx}        
\usepackage{multicol}        
\usepackage[bottom]{footmisc}

\usepackage{newtxtext}       %
\usepackage{newtxmath}       

\usepackage{mwe}

\makeindex             

\newcommand{\yaw}{\psi}

\input{preamble}

\begin{document}
\mainmatter

\title{Decentralized Uncertainty-Aware Active Search with a Team of Aerial Robots}
\titlerunning{Decentralized Uncertainty-Aware Active Search with a Team of Aerial Robots}  

\author{Wennie Tabib \and John Stecklein \and Caleb McDowell \and Kshitij Goel \and Felix Jonathan \and
  Abhishek Rathod \and Meghan Kokoski \and Edsel Burkholder \and Brian Wallace \and Luis Ernesto Navarro-Serment \and
  Nikhil Angad Bakshi \and Tejus Gupta \and  Norman Papernick \and  David Guttendorf \and
  Erik E. Kahn \and  Jessica Kasemer \and  Jesse Holdaway \and Jeff Schneider
}

\authorrunning{Tabib et al.} 

\institute{The Robotics Institute, Carnegie Mellon University, Pittsburgh, PA 15213 USA\\
  \{\texttt{wtabib, jsteckle, cmcdowel, kgoel1, fjonatha, arathod2, mkokoski, eburkhol,
    luisn, nabakshi, tejusg, norm, davidg, eekahn, jkasemer, jholdawa, jeff4}
  \}
  \texttt{@andrew.cmu.edu}%
  \vspace{-0.75cm}
}%

%
%
\maketitle

\begin{figure}[H]
    \centering
  \ifthenelse{\equal{\arxivmode}{true}}{
    \includegraphics[width=\linewidth]{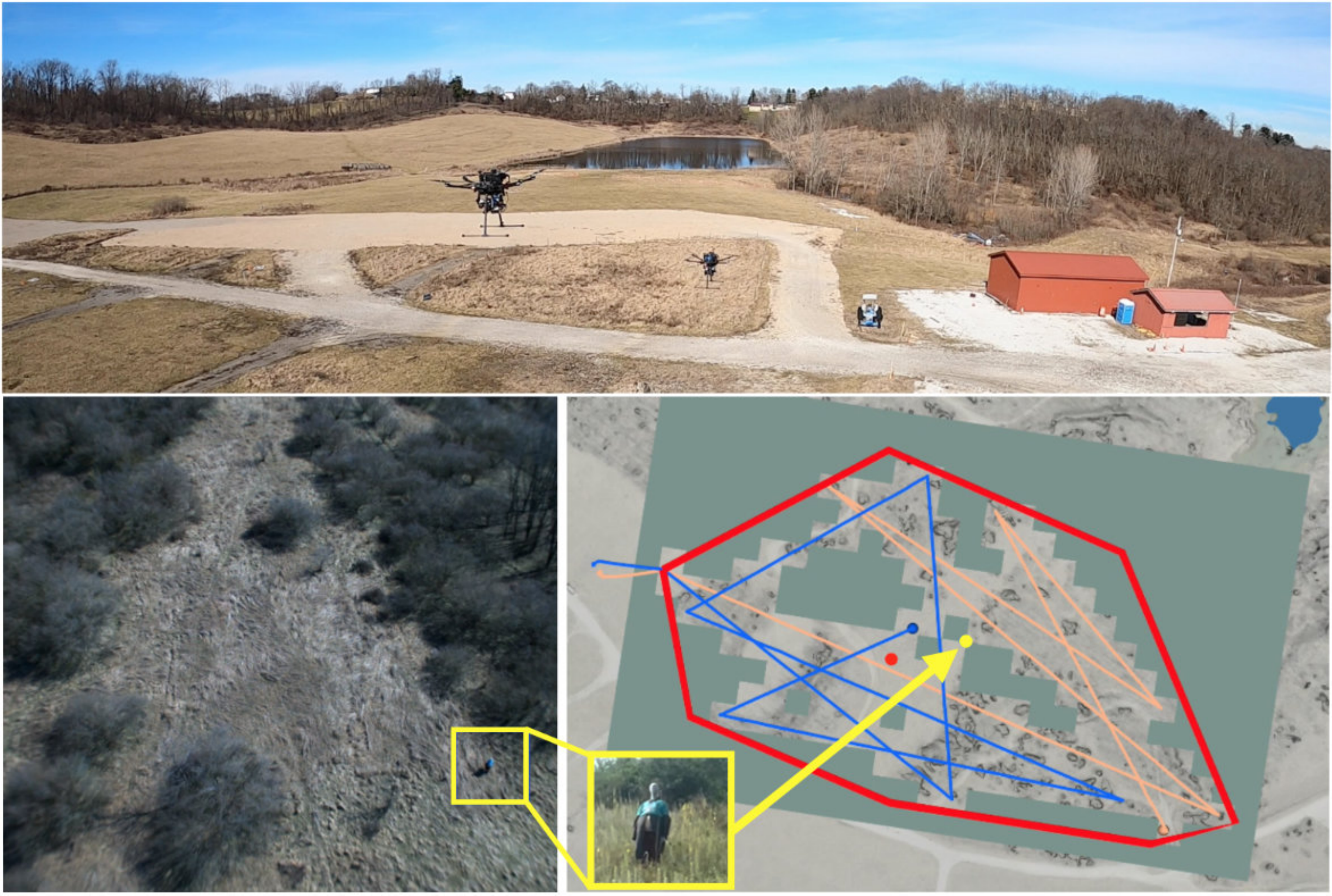}
    }{
    \includegraphics[width=\linewidth]{images/gloryshot5.eps}
    }
    \caption{\label{fig:gloryshot}The decentralized,
        multi-agent team of aerial robots autonomously searches for and
        localizes objects of interest (OOIs) with approximately \SI{3}{\meter}
        error at altitudes up to \SI{60}{\meter}. (top) A team of three
        aerial systems conducts active search. (bottom-left) Onboard view
        of OOI (outlined in yellow) detected in real-time. (bottom-right)
        Location of OOI plotted on shared map generated by the robot team
        with trajectories plotted as lines. The OOI detection is shown as a yellow
        dot. A view of the OOI taken on the ground is inset.
    }
  \vspace{-0.5cm}
\end{figure}

\abstract{Rapid search and rescue is critical to maximizing survival
rates following natural disasters. However, these efforts are
challenged by the need to search large disaster zones, lack of
reliability in the communications infrastructure, and \emph{a priori}
unknown numbers of objects of interest (OOIs), such as injured
survivors. Aerial robots are increasingly being deployed for search
and rescue due to their high mobility, but there remains a gap in
deploying multi-robot autonomous aerial systems for methodical search
of large environments. Prior works have relied on preprogrammed paths
from human operators or are evaluated only in simulation. We bridge
these gaps in the state of the art by developing and demonstrating a
decentralized active search system, which biases its trajectories to
take additional views of uncertain OOIs. The methodology leverages
stochasticity for rapid coverage in communication denied scenarios.
When communications are available, robots share poses, goals, and OOI
information to accelerate the rate of search. Detections from
multiple images and vehicles are fused to provide a mean and
covariance for each OOI location. Extensive simulations and hardware
experiments in Bloomingdale, OH, are conducted to validate the
approach. The results demonstrate the active search approach
outperforms greedy coverage-based planning in communication-denied
scenarios while maintaining comparable performance in
communication-enabled scenarios. The results also demonstrate
the ability to detect and localize all \emph{a priori} unknown OOIs
with a mean error of approximately \SI{3}{\meter} at flight altitudes
between \SIrange{50}{60}{\meter}.
}

\section{Supplementary Materials}
Videos of the experiments are available at:
\begin{itemize}
  \item Two robot search with OOIs: \url{https://youtu.be/qhJS2JhdbAE};
  \item Three robot search with Apriltags: \url{https://youtu.be/xgLnS2IFCQQ}; and
  \item Two robot search without targets: \url{https://youtu.be/lzh8Ml34enw}.
\end{itemize}
Open-source software is available at: \url{https://github.com/rislab/guts-sandbox}.

\section{Introduction}
Rapid emergency response is key to maximizing the survival rate
following a disaster. Rescuing a victim within the first 24 hours
yields a survival rate of 90\%, which drops precipitously to 5-10\%
after 72 hours~\cite{hakami2013application}. Due to their speed,
agility, and maneuverability in challenging three dimensional
environments, unmanned aerial systems are increasingly being deployed
to facilitate search and rescue~\cite{Ashour2023}.  Because the number
of victims may be \emph{a priori} unknown, automated methods are
needed to systematically and methodically cover the disaster area. An
additional challenge arises when communication networks fail, which
may occur during severe natural disasters.  The problem this
work seeks to address is how to autonomously and rapidly search an
area to discover objects of interest (OOIs), such as injured persons,
while remaining robust to communication dropouts.

There have been several recent works in autonomy for search and
environment monitoring that are related to our work.
\citet{stache2023adaptive} develop an environment monitoring system,
which uses a Gaussian Process as a decision function, to modulate an
aerial robot's altitude according to the accuracy of semantic
segmentation. The objective is to maximize the classification accuracy
of objects in the images. While the work is evaluated in simulation,
it is not deployed in real world experiments. In contrast, our active
search methodology is demonstrated with a team of up to three aerial
robots. \citet{horyna2023decentralized} leverage multi-agent flocking
behaviors to increase the reliability of OOI detection. When an OOI is
detected, one of the agents may separate to confirm
the detection. In contrast, the loss function in this work enables
reflies over uncertain OOIs while maximizing coverage of the
space.

\textbf{Contributions:} We extend prior work in active search
by~\citet{guts} and provide the following contributions. First, we
modify the loss function to incorporate goals from other robots when
planning the next best action and provide analysis for a multi-agent
aerial team both in simulation and real-world experiments.  We also
analyze the effects of communication in real-world hardware trials and
report the OOI localization accuracy. Finally, we provide an efficient
C++ implementation of the active search algorithm with Python bindings
for ease of prototyping and deployment to hardware.


\section{Technical Approach\label{sec:methodology}}
\begin{figure}[t!]
  \centering
  \includegraphics[width=\linewidth,trim=70 105 10 150,clip]{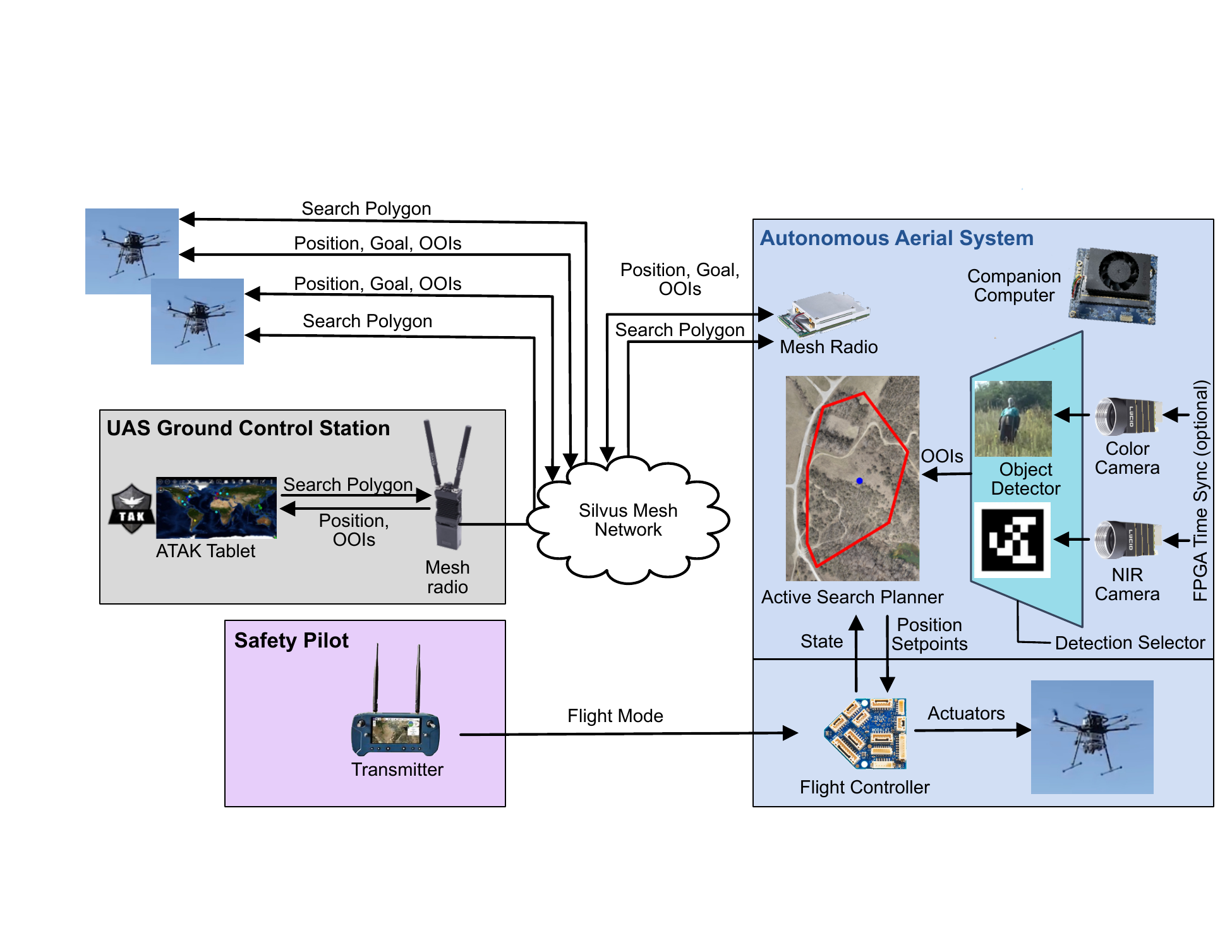}
  \caption{\label{fig:system_architecture}  System diagram for the active search approach. An operator uses
    the Android Team Awareness Kit (ATAK) app on a tablet to draw a convex
    polygon of an area for the aerial systems to search. The polygon is
    sent to one or more robots over a Silvus mesh radio network. Safety pilots
    launch the vehicles. All search operations are conducted without human
    intervention. The robot receives state information from the flight
    controller and camera images are processed to localize
    OOIs on the ground below. The planner sends position
    setpoints to the flight controller, which are used to send actuator
    commands to the motors. When communications are enabled, the robot
    transmits position, target, and goal information to other robots.
    When the battery is depleted, the robots
    return to their takeoff locations and the safety pilots land their
    vehicles.}
  \vspace{-0.5cm}
\end{figure}

\paragraph{Notation}
Lowercase boldface symbols represent column vectors
(e.g., $\mathbf{b}$). The $i$th entry of a vector $\mathbf{b}$ is
denoted as $b_i$. Uppercase bold letters (e.g., $\mathbf{B}$)
represent matrices or sets.  Sets and matrices may have the subscript
$\mathbf{B}_{a:b}$, which means that the matrix or set is composed of
data (e.g., row vectors or scalars) from timestamp $a$ through
timestamp $b$, inclusive. The transpose of a matrix $\mathbf{B}$ is
denoted as $\mathbf{B}^{\top}$.  A square matrix with nonzero entries
$\mathbf{b}$ along the diagonal is represented as diag($\mathbf{b}$).
The $\ell_2$-norm of a vector $\mathbf{b}$ is written as $\lVert
\mathbf{b} \rVert_2$. Where it is important to denote that a
particular value is maintained by robot $j$, it is written as a
superscript (i.e., $\mathbf{b}^j$).

\subsection{Active Search\label{ssec:active_search}}
This section details the active search methodology
for a single robot (i.e., the communication-denied
scenario). It is extended to the communication-enabled case in the
next section.
\Cref{fig:system_architecture} provides a system diagram of the approach.
Let the environment be represented as
a 2D discrete grid with dimension $M = M_{w} \times M_{h}$,
which may be flattened into a vector, $\pmb{\beta} \in
\mathbb{R}^{M}$. $\mathbf{x}_i \in \mathbb{R}^{1\times M}$ is a
one-hot sensing row vector. The aerial system sensing action model
consists of all the cells along the straight line between a start and
end point.
Therefore, an action may be represented using multiple sensing row
vectors.
Robot positions and OOI detection locations are
encoded in $\mathbf{x}_i$ with 1. All other entries are 0.  Each robot
also maintains a scalar value $y_i$, which represents the output of
the object detector as well as the OOI confidence, $c_i$.

The data $\mathbf{D}_{1:i} = \{(\mathbf{x}_1, y_1), \ldots, (\mathbf{x}_i,
y_i)\}$ consists of all sensor row vectors and observations up to
timestep $i$. The sensor row vectors are vertically stacked to create
a matrix $\mathbf{X}_{1:i}$ and observations $y_i$ are also vertically
stacked to create a column vector $\mathbf{y}_{1:i}$. For example, if
$\mathbf{D}_{1:3} = \{(\mathbf{x}_1, y_1), (\mathbf{x}_2, y_2),
(\mathbf{y}_3, y_3)\}$, then $\mathbf{X}_{1:3} = \begin{bmatrix}
\mathbf{x}_1 & \mathbf{x}_2 & \mathbf{x}_3 \end{bmatrix}^{\top}$ and
$\mathbf{y}_{1:3} = \begin{bmatrix} y_1 & y_2 &
y_3 \end{bmatrix}^{\top}$. Noise is modeled with a diagonal matrix
$\pmb{\Sigma}_{1:i} = \mbox{diag}\left( \begin{bmatrix} \sigma_1^{2} &
\ldots & \sigma_i^{2} \end{bmatrix}^{\top} \right)$, where
$\sigma_i^{2} = \frac{1}{c_i}$.

Expectation Maximization is used to estimate the posterior
distribution of $\pmb{\beta}$ given data $\mathbf{D}_{1:i}$,
$p(\pmb{\beta}|\mathbf{D}_{1:i}, \pmb{\Gamma}) = \mathcal{N}(\pmb{\mu},
\mathbf{V})$, where $\pmb{\mu}$ and $\mathbf{V}$ are defined as
\begin{align*}
\mathbf{V} &= (\pmb{\Gamma}^{-1} + \mathbf{X}_{1:i}^{\top} \pmb{\Sigma}_{1:i} \mathbf{X}_{1:i})^{-1}\\
\pmb{\mu} &= \mathbf{V}\mathbf{X}_{1:i}^{\top}\pmb{\Sigma}_{1:i}\mathbf{y}_{1:i}
\end{align*}
and where $\pmb{\Gamma} \in \mbox{diag}\left(\begin{bmatrix} \gamma_1
& \ldots & \gamma_M \end{bmatrix}^{\top} \right)$ are the hidden
variables~\cite{guts,nats}. It should be noted that $\pmb{\Gamma} \in
\mathbb{R}^{M \times M}$ and $\pmb{\mu} \in \mathbb{R}^M$.

The Maximization step maximizes the likelihood of
$p(\mathbf{y}_{1:i}|\pmb{\Gamma}, \mathbf{X}_{1:i})$ such that the
responsibilities $\gamma_m$, where $m \in [1, \ldots, M]$, may be
calculated as:
\begin{align*}
  \gamma_m = ([\mathbf{V}]_{mm} + [\mathbf{\mu}]_m^2 + 2b_m) / (1+2a_m).
\end{align*}
$a_m = 0.1$ and $b_m = 1$ in keeping with~\cite{nats,guts}.  After
these operations are complete, the robot samples from the posterior
$\tilde{\pmb{\beta}} \sim p(\pmb{\beta}|\mathbf{D}_{1:i})$.

To select the next sensing action $\mathbf{X}_{i+1:n} = \begin{bmatrix}
\mathbf{x}_{i+1} & \ldots & \mathbf{x}_{n} \end{bmatrix}^{\top}$,
which lies along a straight line trajectory to a candidate goal location,
each agent minimizes the loss function~$\mathcal{L}(\tilde{\pmb{\beta}}, \mathbf{D}_{1:i}, \mathbf{X}_{i+1:n})$
\begin{align}
  \mathcal{L}(\tilde{\pmb{\beta}}, \mathbf{D}_{1:i}, \mathbf{X}_{i+1:n}) &= \lVert \tilde{\pmb{\beta}}-\hat{\pmb{\beta}}\rVert_2 + \lambda \mbox{I}(\tilde{\pmb{\beta}},\hat{\pmb{\beta}})\label{eq:5}
\end{align}
where $I(\tilde{\pmb{\beta}}, \hat{\pmb{\beta}})$ is an
indicator function (defined in~\cref{eq:indicator}), which
encourages the selection of actions that may detect targets
outside the most confident set in the sample from the belief,
$\hat{\pmb{\beta}}$.  $\hat{\pmb{\beta}}$ is calculated as
\begin{align}
  \hat{\pmb{\beta}} &= \mathbf{H}_i\mathbf{y}_{1:i} + \mathbf{H}_n\mathbf{X}_{i+1:n}\tilde{\pmb{\beta}}\label{eq:4}\\
  \begin{bmatrix}\mathbf{H}_i & \mathbf{H}_n \end{bmatrix} &= \mathbf{\mathbf{S}(\begin{bmatrix}\mathbf{X}_{1:i}^{\top} \pmb{\Sigma}_{1:i} & \mathbf{X}_{i+1:n}^{\top}\pmb{\Sigma}_{i+1:n}\end{bmatrix})}\label{eq:3}\\
  \mathbf{S} &= \mbox{diag}\left(\left(\mathbf{U}_{k,k} \right)^{-1} \right) \label{eq:2}\\
  \mathbf{U} &= (\left(\mathbf{X}^{\top}_{1:i} \pmb{\Sigma}_i \mathbf{X}_{1:i} + \pmb{\Gamma}^{-1} \right) \notag\\
  &~~~~+ \mathbf{X}_{i+1:n}^{\top} \pmb{\Sigma}_{i+1:n} \mathbf{X}_{i+1:n} ) \odot \mathbb{I} \label{eq:1}
\end{align}
$\odot$ denotes elementwise multiplication and $\mathbb{I}$ represents
the identity matrix. In~\cref{eq:2}, the notation $\mbox{diag}\left(
(\mathbf{U}_{k,k})^{-1} \right)$ represents extracting the diagonal
entries of the matrix $\mathbf{U}$, taking the inverse of these
entries, and converting the column vector into a diagonal matrix.

$I(\tilde{\pmb{\beta}}, \hat{\pmb{\beta}})$ is determined by finding half
the maximum value of $\hat{\pmb{\beta}}$ and $\tilde{\pmb{\beta}}$,
checking if the corresponding value of $\hat{\beta}_k$ and
$\tilde{\beta}_k$ is larger, respectively, and then rounding to the
nearest integer (0 or 1).  If all elements of $\mathbf{\hat{a}}$ match
all elements of $\mathbf{\tilde{a}}$, then the indicator function
specified in~\cref{eq:indicator} returns a 0 and 1, otherwise.

\begin{align}
\hat{a}_k &= \round{\hat{\beta}_k > \max(\hat{\pmb{\beta}})/2}\label{eq:6}\\
\tilde{a}_k &= \round{\tilde{\beta}_k > \max(\tilde{\pmb{\beta}})/2}\label{eq:7}\\
I(\tilde{\pmb{\beta}}, \hat{\pmb{\beta}}) &= \begin{cases}
  0, &\text{if } \hat{a}_k = \tilde{a}_k,~\forall k\\
  1, & \text{otherwise}
\end{cases}\label{eq:indicator}
\end{align}
$\lambda = 0.01$ in keeping
with~\cite{guts,bakshiStealthyTerrainAwareMultiAgent2023a}.

\subsection{Decentralized Multirobot Planning\label{ssec:decentralized_planning}}
This section details how the active search loss function changes when
multiple robots share position, goal, and OOI information.  Positions,
goals, and tracks from other robots are incorporated through updates
to the $\mathbf{X}_{1:i}$ and $\mathbf{y}_{1:i}$ variables.  We will
consider what happens to these variables when robot $j$ receives
information from robot $k$.

When robot $j$ receives location information from robot $k$,
robot $j$ generates a sensor row vector $\mathbf{x}_{i+1}^k$ and
observation scalar $y_{i+1}^k$ and appends it to the variables
$\mathbf{X}_{1:i}$ and $\mathbf{y}_{1:i}$ as in the following
equations:
\begin{align*}
  \mathbf{X}_{1:i+1} &= \begin{bmatrix} \mathbf{x}_1^j & \ldots & \mathbf{x}_i^j & \mathbf{x}_{i+1}^k \end{bmatrix}^{\top}\\
  \mathbf{y}_{1:i+1} &= \begin{bmatrix} y_1^j & \ldots & y_i^j & y_{i+1}^k \end{bmatrix}^{\top}.
\end{align*}
The same update is used when robot $j$ receives an OOI detection from
robot $k$.

When goals are transmitted, the update is slightly different because
there are multiple cells that robot $k$ traverses. If robot $j$ has
received information that robot $k$ is at the position encoded in
$\mathbf{x}^k_{i+1}$, and the next sensing action for robot $k$ may be
specified as $\mathbf{X}_{i+2:p}^k = \begin{bmatrix} \mathbf{x}_{i+2}^k & \ldots & \mathbf{x}_{p}^k \end{bmatrix}^{\top}$,
then robot $j$ appends to the variables $\mathbf{X}_{1:i+1}$
and $\mathbf{y}_{1:i+1}$ in the following way:
\begin{align*}
  \mathbf{X}_{1:p} &= \begin{bmatrix} \mathbf{x}_1^j & \ldots & \mathbf{x}_i^j & \mathbf{x}_{i+1}^k & \mathbf{x}_{i+2}^k & \ldots & \mathbf{x}_{p}^k\end{bmatrix}^{\top}\\
  \mathbf{y}_{1:p} &= \begin{bmatrix} y_1^j & \ldots & y_i^j & y_{i+1}^k & y_{i+2}^k & \ldots & y_{p}^k \end{bmatrix}^{\top}.
\end{align*}
$\mathbf{D}^j_{1:p} = (\mathbf{X}_{1:p}, \mathbf{y}_{1:p})$ is used
in~\cref{eq:5} in place of $\mathbf{D}_{1:i}$ since it contains all
the information available to robot $j$.  Formulating
$\mathbf{D}_{1:p}^j$ such that it includes goal information from other
robots ensures that robot $j$ will avoid visiting the same
locations.

\subsection{Trajectory Design and Tracking\label{sec:traj}}
Because the hexarotor aerial system used in this work is
differentially flat~\cite{mellinger2011}, single axis trajectories for
$x$, $y$, $z$ position as well as heading ($\yaw$) may be formulated
for the robot to track between waypoints. We leverage the single-axis
trajectory generation solution of~\citet{mueller} because it is
computationally efficient for size, weight, and power constrained
aerial systems. It also enables the time-parameterized calculation of
position setpoints, which enables re-planning in flight using all
available information up to $\Delta t$ seconds before reaching the
next waypoint.
The roots of the derivative of a given polynomial are the times at
which the polynomial achieves a critical point, which can be used to
ensure the vehicle limits are not exceeded.

\subsection{Aerial System Hardware Configuration\label{sec:aerial_system}}
The aerial system platform is an Inspired Flight 1200 with dimensions
\SI{1.4}{\meter} $\times$ \SI{1.3}{\meter} $\times$ \SI{0.7}{\meter} (see~\cref{fig:gloryshot}).
It uses a
ModalAI\footnote{\url{https://docs.px4.io/main/en/flight_controller/modalai_fc_v1.html}}
flight controller running a closed-source
fork of the ModalAI firmware tag
v1.11.3-0.2.3\footnote{\url{https://github.com/modalai/px4-firmware/releases/tag/v1.11.3-0.2.3}},
which is a modified version of the PX4 Autopilot firmware.

Two sensing payloads are developed for the systems.
The first sensing payload consists of a global shutter Lucid Vision
Phoenix NIR camera, Lucid Vision Phoenix RGB camera, and Teledyne
Calibir GX camera.  The companion computer is a NVIDIA Jetson
AGX Xavier.  Silvus Technologies' StreamCaster 4240 mesh
radios enable communications between aerial systems and operators via
a mobile ad hoc network.  The mesh network enables the aerial systems
to automatically and dynamically route information between one another
and intermediate participating radios (e.g., ground control stations,
participating operators, and dedicated relay points).

The second sending payload upgrades the companion computer to
a NVIDIA Orin NX with \SI{16}{\giga\byte} RAM and consists of only
the RGB camera and Silvus radio.

The NIR camera is used for Apriltag detection experiments and
the RGB camera is used for OOI detection. All cameras face at
approximately $45^\circ$ to the ground and communicate with the
companion computer via GigE.  A tablet running geospatial mapping
software called the Android Team Awareness Kit (ATAK) is on the
network and transmits zone information to the aerial systems. The
aerial systems transmit their poses and object detections to be
visualized on the tablet.

\subsection{Cross-Platform Fusion\label{sec:cpf}}
During flight, a single OOI may be detected across multiple
images and platforms. To accurately localize the OOI, multiple
observations of the object are fused into a single estimate, or track. The
process of merging the detections into a single track estimate is
called cross-platform fusion (CPF).  There are two stages to CPF: the
first step associates the new detection with currently fused
tracks and the second step fuses the tracks. The association
step leverages the Euclidean distance from the mean position
of the track, which works well when the OOIs are assumed to be
stationary.  Once the distance cost is calculated, we use the
Hungarian algorithm~\cite{kuhn1955hungarian} to find the best
association with both fused and unfused tracks. If the
distance exceeds a pre-defined threshold to any known track, a new track is
created. Finally, we iterate through the fused tracks to determine
whether or not they should be combined with other fused tracks.  Once
two tracks have been merged, they will stay merged until a greater
distance threshold is reached. Fused state is calculated using the
inverse covariance intersection~\cite{cfp}, which yields a mean
and covariance.


\section{Experiment Design and Results\label{sec:experiments}}
\begin{figure}
  \centering
  \ifthenelse{\equal{\arxivmode}{true}}{
    \subfloat[\label{sfig:wingtraone}WingtraOne]{\includegraphics[width=0.29\linewidth,trim=55 0 150 0,clip]{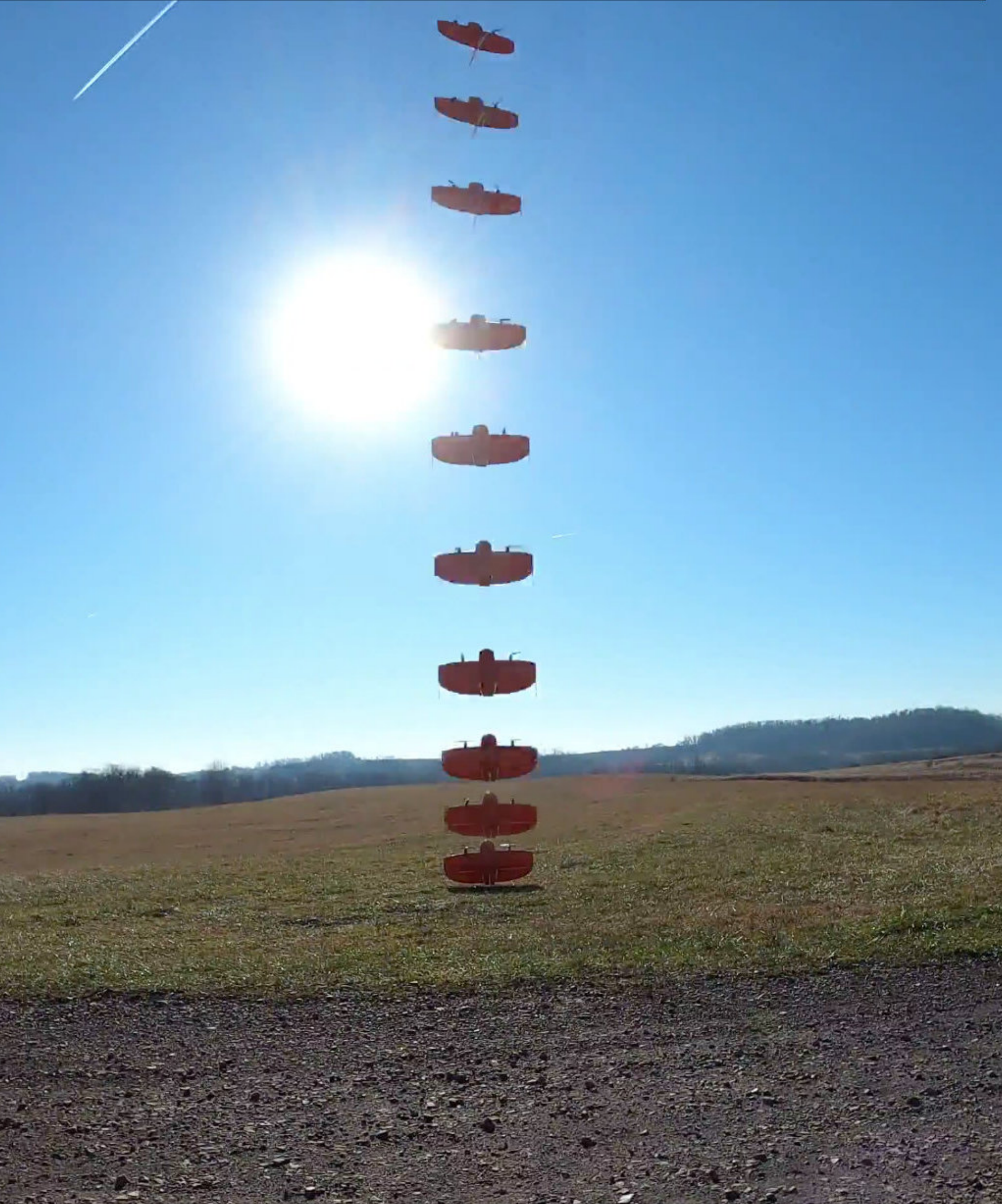}}\hfill%
  }{
    \subfloat[\label{sfig:wingtraone}WingtraOne]{\includegraphics[width=0.29\linewidth,trim=55 0 150 0,clip]{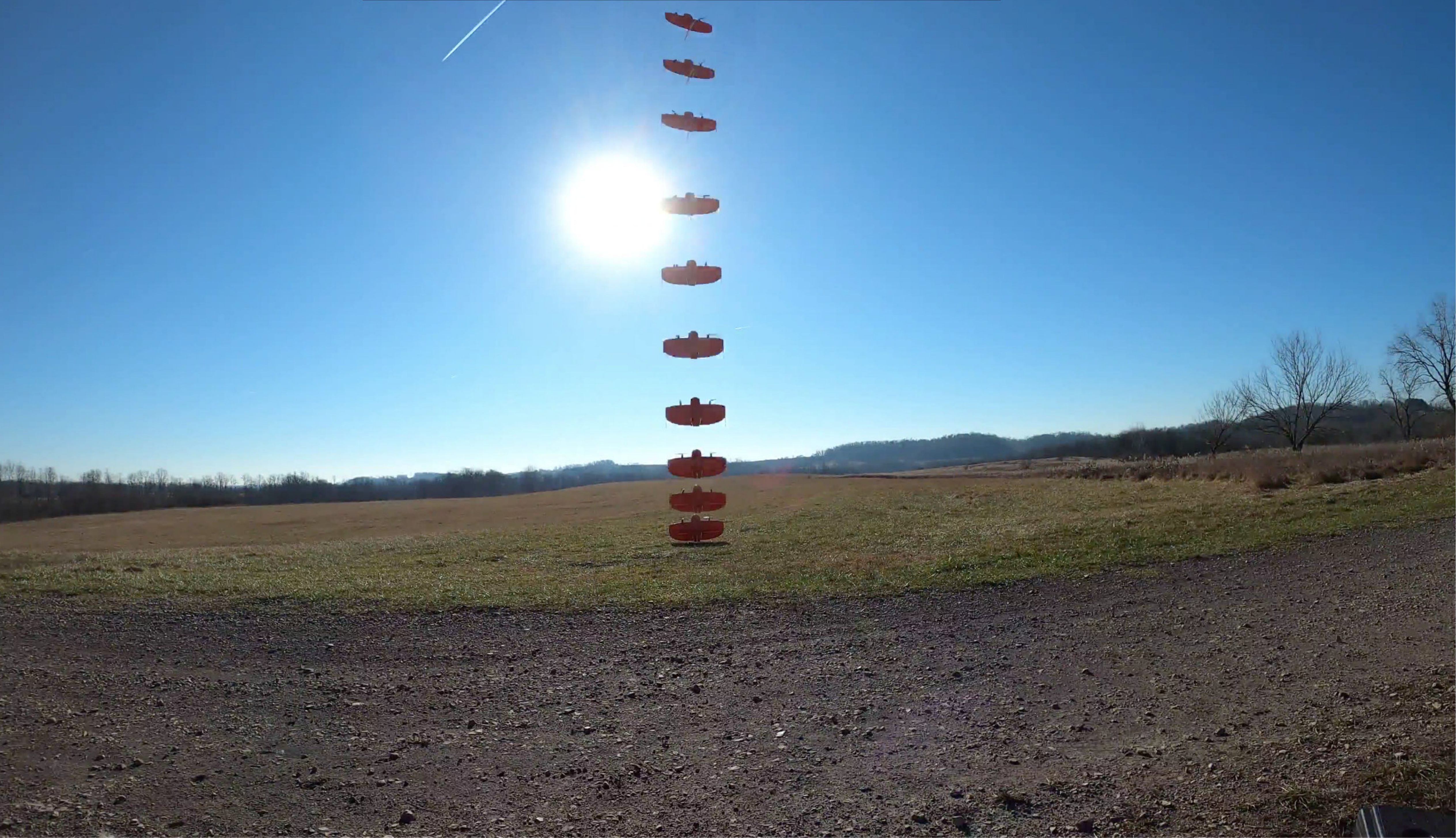}}\hfill%
  }%
  \ifthenelse{\equal{\arxivmode}{true}}{
    \subfloat[\label{sfig:pix4dmap} Reconstruction]{\includegraphics[width=0.3\linewidth,trim=100 20 50 0,clip]{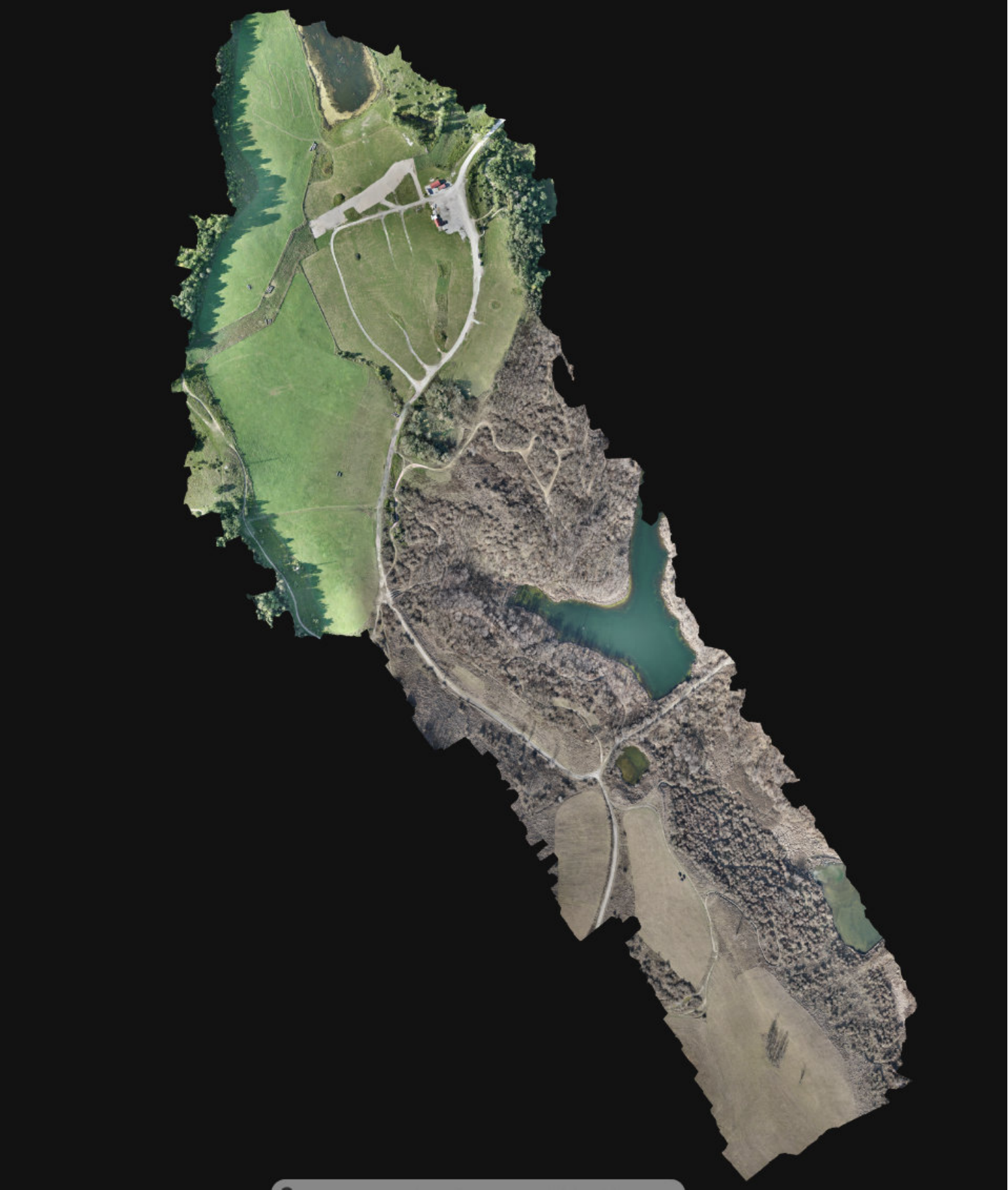}}\hfill%
  }{
    \subfloat[\label{sfig:pix4dmap} Reconstruction]{\includegraphics[width=0.3\linewidth,trim=100 20 50 0,clip]{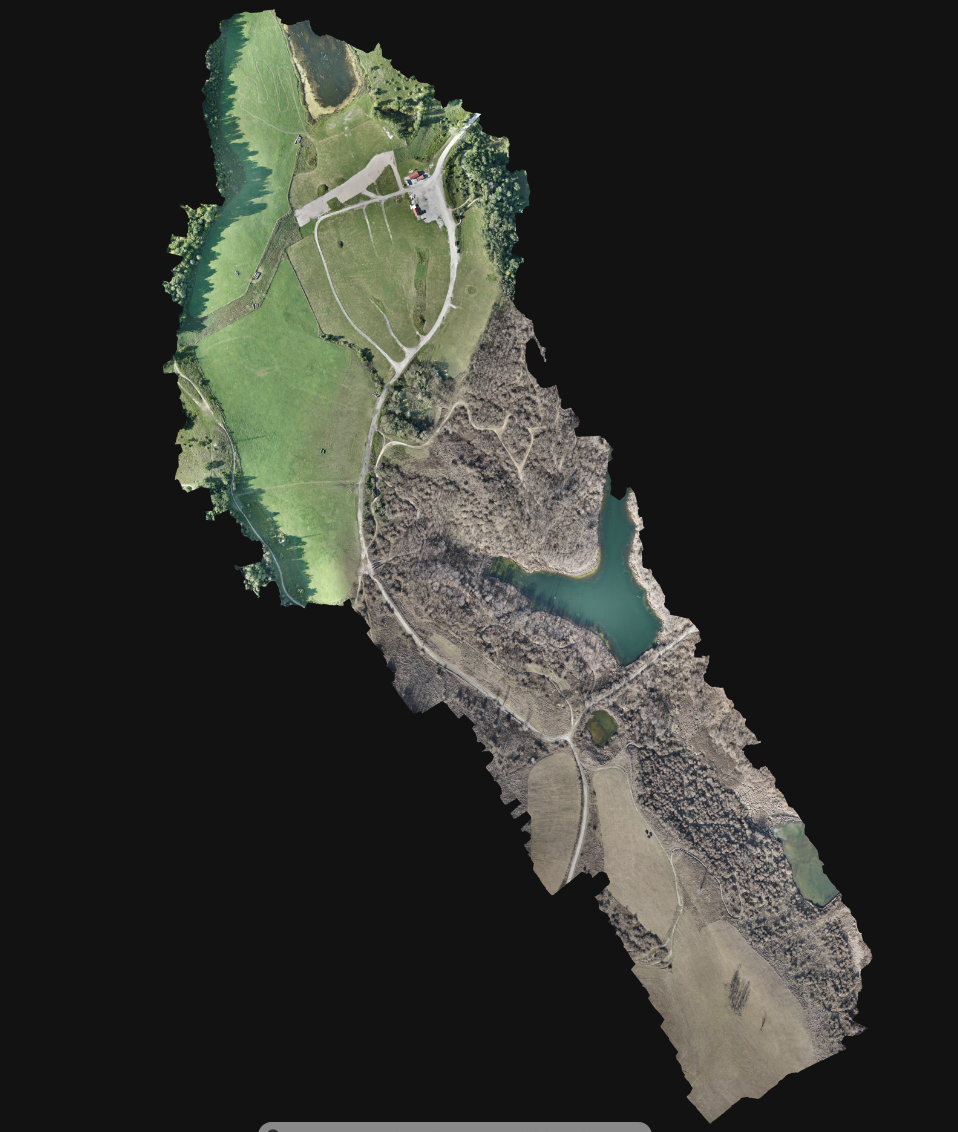}}\hfill%
  }%
  \ifthenelse{\equal{\arxivmode}{true}}{%
    \subfloat[\label{sfig:wingtratopdown} View from the Wingtra One]{\includegraphics[width=0.385\linewidth,trim=0 0 0 0,clip]{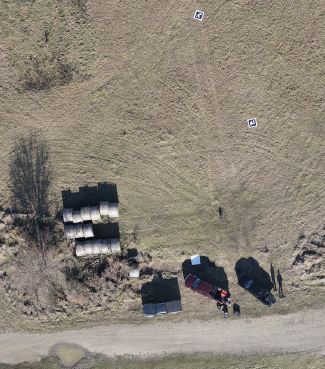}}\hfill%
    }{
    \subfloat[\label{sfig:wingtratopdown} View from the Wingtra One]{\includegraphics[width=0.385\linewidth,trim=0 0 0 0,clip]{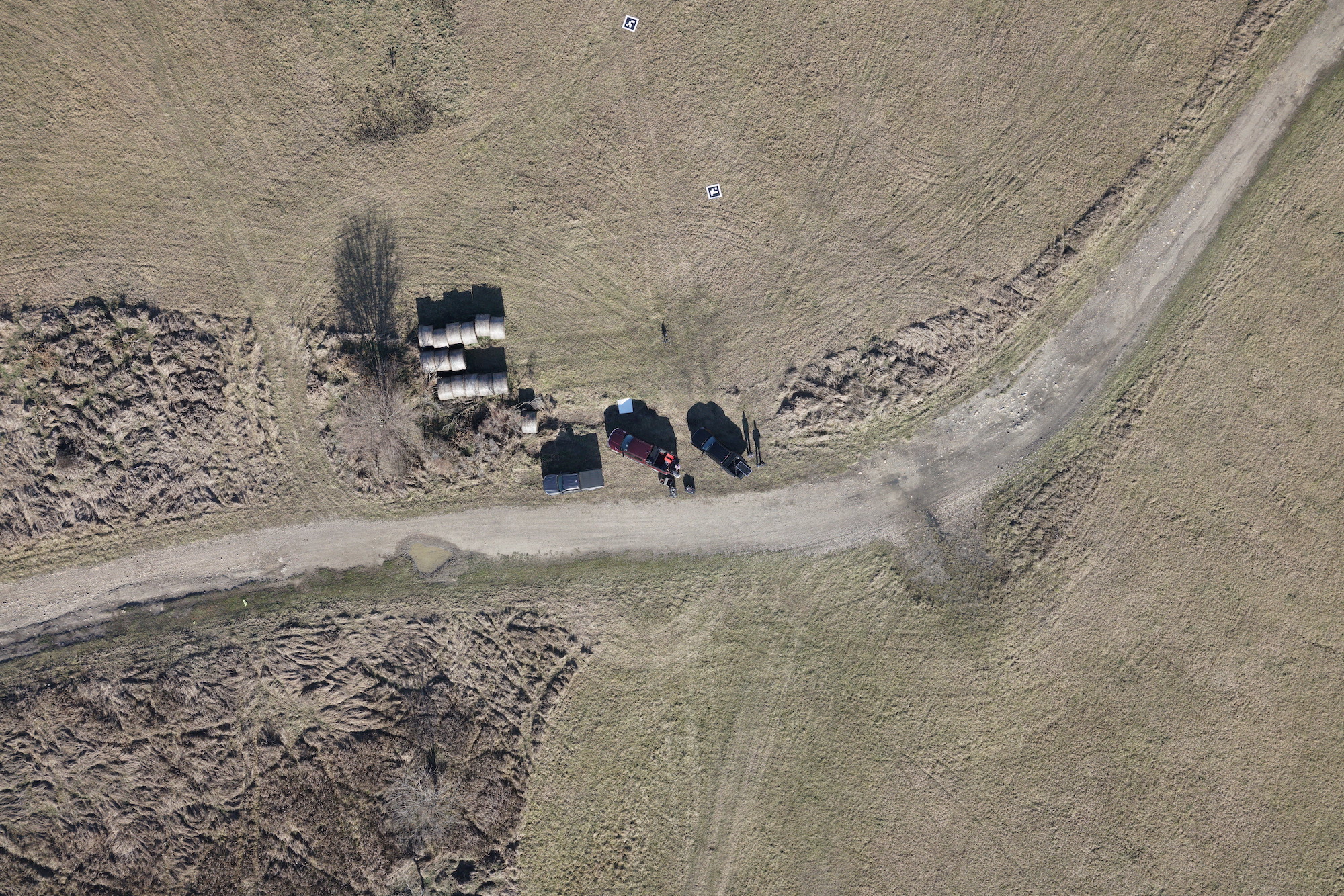}}\hfill%
    }
  \caption[]{\label{fig:wingtra}The~\protect\subref{sfig:wingtraone}
    WingtraOne\footnote{\url{https://wingtra.com}} VTOL is used to collect
    images of the test site and produce a~\protect\subref{sfig:pix4dmap}
    high-resolution, geo-registered point cloud and mesh of the
    environment using the Pix4DMatic\footnote{\url{https://www.pix4d.com}}
    photogrammetry software.~\protect\subref{sfig:wingtratopdown}
    illustrates a view from the WingtraOne. This image is the test set up
    for the results shown in~\cref{fig:hardware_tracks}.}
  \ifthenelse{\equal{\arxivmode}{true}}{
    \subfloat[\label{sfig:comms_disabled_no_tracks}Communication disabled]{\includegraphics{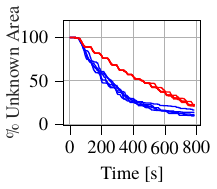}}%
  }{
    \subfloat[\label{sfig:comms_disabled_no_tracks}Communication disabled]{\input{figures/simulation/simulation_comms_disabled_2023-12-05-percentage.tex}} \hfill%
  }
  \ifthenelse{\equal{\arxivmode}{true}}{
    \subfloat[\label{sfig:comms_enabled_no_tracks}Communication enabled]{\includegraphics{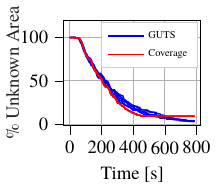}}%
  }{
    \subfloat[\label{sfig:comms_enabled_no_tracks}Communication enabled]{\input{figures/simulation/simulation_comms_enabled_2023-12-05-percentage.tex}} \hfill%
  }
  \subfloat[Coverage]{\includegraphics[width=0.17\textwidth]{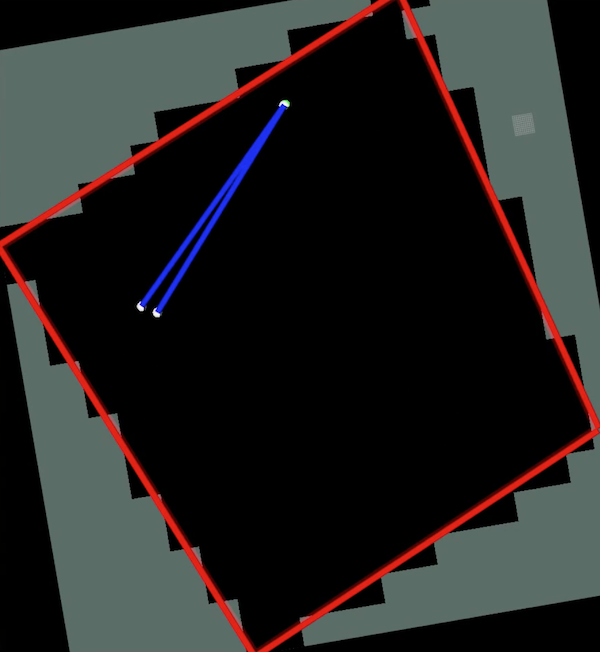}}\hfill
  \subfloat[GUTS]{\includegraphics[width=0.1665\textwidth,trim=40 20 30 20,clip]{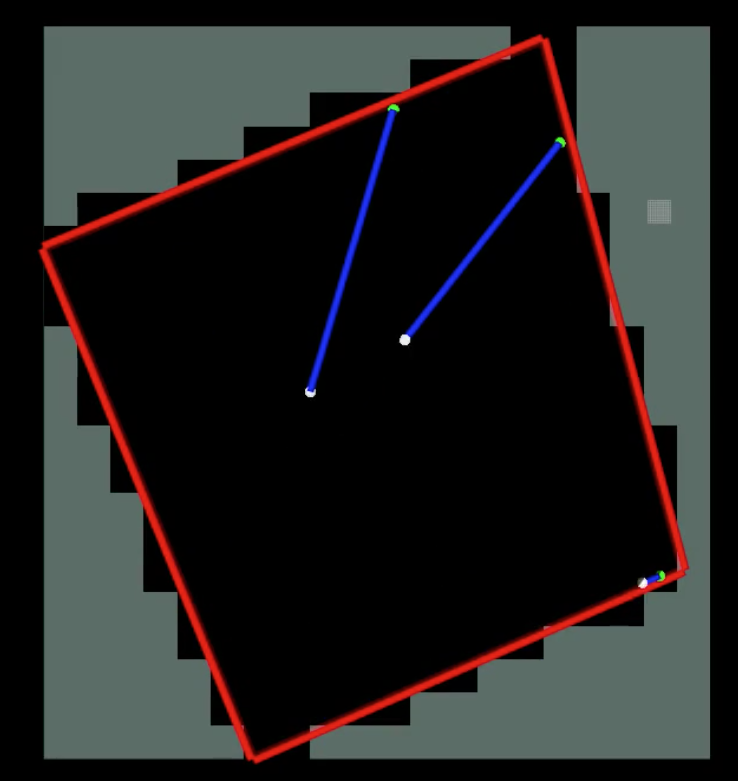}}\hfill
  \caption{\label{fig:comms_no_tracks}Simulation results of three
    robots with and without communication enabled between robots. The
    results highlight that the stochasticity of the GUTS planner provides
    better coverage when~\protect\subref{sfig:comms_disabled_no_tracks}
    communication is disabled.~\protect\subref{sfig:comms_enabled_no_tracks} The GUTS
    planner suffers only a slight performance decrease as compared to the
    coverage planner when communication is enabled. Each approach is run
    five times for \SI{800}{\second} for a total of 20 trials.}
  \ifthenelse{\equal{\arxivmode}{true}}{
    \subfloat[\label{sfig:tracks_comms_enabled}Communication disabled]{\includegraphics{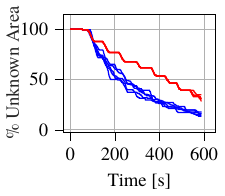}}\hfill%
  }{
    \subfloat[\label{sfig:tracks_comms_enabled}Communication disabled]{\input{figures/simulation/231218_simulation_tracks_comms_denied.tex}}\hfill%
  }
  \ifthenelse{\equal{\arxivmode}{true}}{
    \subfloat[\label{sfig:tracks_comms_disabled}Communication enabled]{\includegraphics{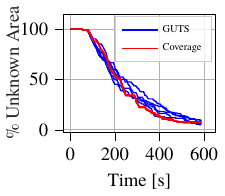}}\hfill%
  }{
    \subfloat[\label{sfig:tracks_comms_disabled}Communication enabled]{\input{figures/simulation/231217_comms_enabled_tracks.tex}}\hfill%
  }
  \subfloat[\label{sfig:tracks_enabled_sim_coverage}Coverage]{\includegraphics[width=0.165\textwidth]{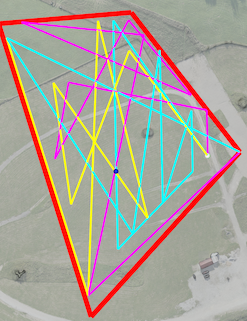}}\hfill%
  \subfloat[\label{sfig:tracks_enabled_sim_nats}GUTS]{\includegraphics[width=0.164\textwidth]{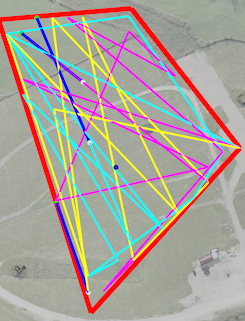}}\hfill%
  \caption{\label{fig:sim_tracks}Simulation results of active search
    with three robots and one OOI (i.e., blue dot
    in~\protect\subref{sfig:tracks_enabled_sim_coverage}
    and~\protect\subref{sfig:tracks_enabled_sim_nats}). The robot
    trajectories are shown in magenta, yellow, and cyan in both figures.
    The blue trajectories are the ones currently being executed in the
    simulation. Similar to~\cref{fig:comms_no_tracks}, the GUTS approach outperforms
    the coverage approach when communications are disabled.
  }
\end{figure}

\begin{figure}[t!]
  \ifthenelse{\equal{\arxivmode}{true}}{
  \subfloat[\label{sfig:target_one_certain}1 target, $c$=0.5, 38 views]{\includegraphics{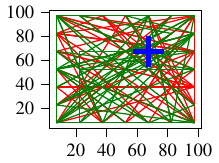}}\hfill%
  }{
  \subfloat[\label{sfig:target_one_certain}1 target, $c$=0.5, 38 views]{\input{figures/sigma_squared/one_target_noise_0.5.tex}}\hfill%
  }
  \ifthenelse{\equal{\arxivmode}{true}}{
    \subfloat[\label{sfig:target_one_less}$c$=0.2, 54 views]{\includegraphics{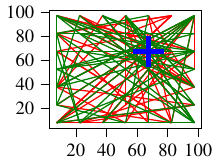}}\hfill%
  }{
    \subfloat[\label{sfig:target_one_less}$c$=0.2, 54 views]{\input{figures/sigma_squared/one_target_noise_0.2.tex}}\hfill%
  }
  \ifthenelse{\equal{\arxivmode}{true}}{
    \subfloat[\label{sfig:target_one_least}$c$=0.005, 83 views]{\includegraphics{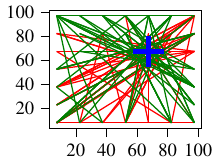}}\hfill%
  }{
    \subfloat[\label{sfig:target_one_least}$c$=0.005, 83 views]{\input{figures/sigma_squared/one_target_noise_0.005.tex}}\hfill%
  }
  \ifthenelse{\equal{\arxivmode}{true}}{
    \subfloat[\label{sfig:target_two_certain}2 targets, $c$=0.5, 73 views]{\includegraphics{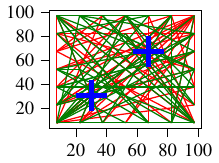}}\hfill%
  }{
    \subfloat[\label{sfig:target_two_certain}2 targets, $c$=0.5, 73 views]{\input{figures/sigma_squared/two_target_noise_0.5.tex}}\hfill%
  }
  \ifthenelse{\equal{\arxivmode}{true}}{
    \subfloat[\label{sfig:target_two_less}$c$=0.2, 99 views]{\includegraphics{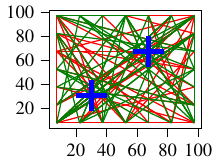}}\hfill%
  }{
    \subfloat[\label{sfig:target_two_less}$c$=0.2, 99 views]{\input{figures/sigma_squared/two_target_noise_0.2.tex}}\hfill%
  }
  \ifthenelse{\equal{\arxivmode}{true}}{
    \subfloat[\label{sfig:target_two_least}$c$=0.005, 148 views]{\includegraphics{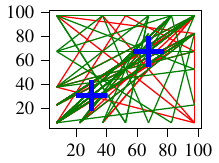}}\hfill%
  }{
    \subfloat[\label{sfig:target_two_least}$c$=0.005, 148 views]{\input{figures/sigma_squared/two_target_noise_0.005.tex}}\hfill%
  }
    \caption{Simulation results that provide qualitative and quantitative examples of the effect of
      varying $c$ for a team of two robots.  The OOI locations are shown as
      blue crosses.  The
      trajectory for robot 1 is shown in red and the trajectory for robot 2
      is shown in green.  When the confidence is high (e.g.,~\protect\subref{sfig:target_one_certain} and~\protect\subref{sfig:target_two_certain})
      the behavior is more exploratory.
      As the certainty value is decreased (e.g.,~\protect\subref{sfig:target_one_less} and~\protect\subref{sfig:target_two_less},
      one may see that the action selection is increasingly clustered around the targets.
      When $c$ is close to 0.0, the certainty is very low, so the
      planner will select points that obtain additional views of target (i.e.,~\protect\subref{sfig:target_one_least} and~\protect\subref{sfig:target_two_least}).
      The number of times the robot views the target
      is counted and provided in the figure caption. The robots will
      select waypoints to fly over the targets more often as the uncertainty
      increases.\label{fig:toy_tracks}}
    \vspace{-0.5cm}
\end{figure}

\subsection{Simulation Results\label{sec:simulation_results}}
The approach is tested in simulation on a Lenovo Thinkpad X1 with an
11th Gen Intel Core i7-1165G7 \SI{2.8}{\giga\hertz} CPU, 4 cores
(8 hyperthreads), and \SI{16}{\giga\byte} RAM running Ubuntu 20.04.
Because the approach is stochastic, multiple trials
are run to evaluate the performance. The
approach is compared to a deterministic coverage planner that greedily
maximizes visiting unknown cells.  The
results are analyzed by plotting the percent reduction in unknown area
as a function of time.  For each approach
illustrated in~\cref{fig:comms_no_tracks} two cases are considered:
communication disabled and communication enabled.
Our approach, labeled Generalized Uncertainty-Aware Thompson Sampling (GUTS),
is shown in blue and the greedy coverage planner
is shown in red.

In the communication-denied scenario, GUTS outperforms
the coverage-based approach due to the stochasticity. The coverage
approach in the communication enabled case
(\cref{sfig:comms_enabled_no_tracks}) slightly outperforms the GUTS
approach early on; however, towards the end of the run the GUTS
approach slightly outperforms it.
For these simulation experiments, no OOIs were included
to enable a thorough analysis of the percent coverage achieved by each
approach.

\Cref{fig:sim_tracks} provides the same analysis but with one
OOI, which is shown as a blue dot.  The performance is similar
to that shown in~\cref{fig:comms_no_tracks}. Three robots are used in
the simulation and the trajectories for each robot are shown in
magenta, yellow, and cyan
in~\cref{sfig:tracks_enabled_sim_coverage,sfig:tracks_enabled_sim_nats}.
The behavior of the GUTS approach is to first explore the environment
without stopping even if it finds an uncertain target. When the
environment is completely covered, it will refly over areas to obtain
better views of the targets.  This is demonstrated with qualitative
simulation results in~\cref{fig:toy_tracks}.

To evaluate the effect of target uncertainty, we vary the value of
$c$ and quantify the number of times the robot views the
target(s). $c$ is the only parameter varied in the
simulations shown in~\cref{fig:toy_tracks}.  When there is
high certainty, the robot executes more exploratory behavior as shown
in~\cref{sfig:target_one_certain,sfig:target_two_certain}.  The 38
and 73 views, respectively, indicate the number of times the robots
views the target(s). As the certainty is decreased, the number of
views increases.


\subsection{Hardware Results\label{sec:hardware_results}}
To validate the simulations from~\cref{sec:simulation_results}, the
algorithms are deployed to hardware using the aerial vehicles
described in~\cref{sec:aerial_system}. Four analyses are conducted to
validate the simulations.  All hardware tests were conducted in
Bloomingdale, OH, a 1500+ acre site. An aerial view of the areas flown
over during testing is shown in~\cref{sfig:pix4dmap}.

The cell size used is \SI{15}{\meter} $\times$ \SI{15}{\meter}.  The
planners plan continuously and the best action is stored. When time
runs out, the best action is published. Due to our efficient C++
implementation, 100\% of the cells are sampled and evaluated at each
planning round.

\begin{figure}
  \ifthenelse{\equal{\arxivmode}{true}}{
    \subfloat[\label{sfig:hardware_comms_denied} Area visited over time]{\includegraphics{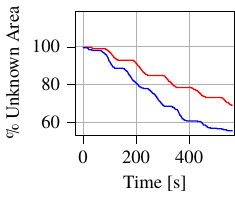}}\hfill %
  }{
    \subfloat[\label{sfig:hardware_comms_denied} Area visited over time]{\input{figures/hardware/renegade_ridge_231018_comms_denied_two_robots_percentage.tex}}\hfill %
  }
  \subfloat[\label{sfig:hardware_comms_denied_coverage_map} Coverage]{\includegraphics[width=0.3\textwidth,trim=0 50 0 50,clip]{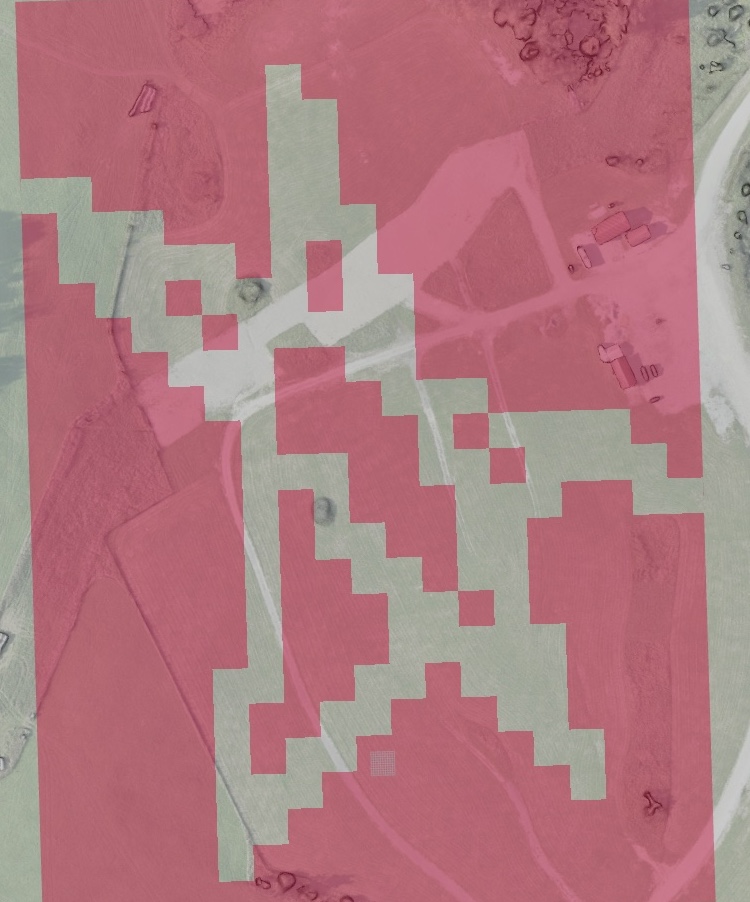}}\hfill%
  \subfloat[\label{sfig:hardware_comms_denied_nats_map}GUTS]{\includegraphics[width=0.31\textwidth,trim=0 50 0 50,clip]{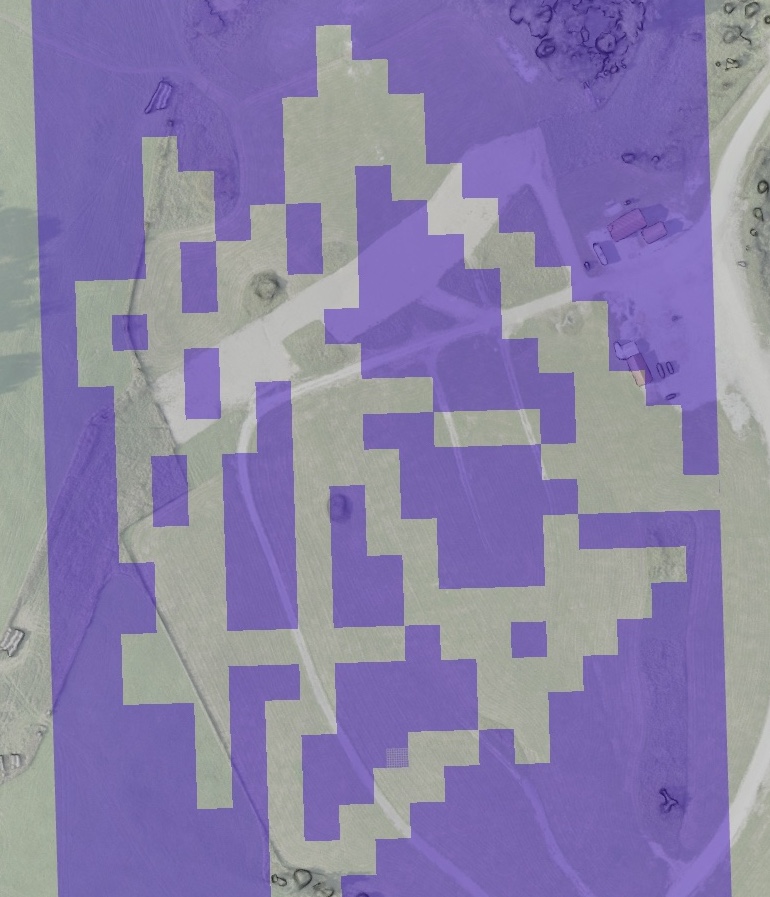}}\hfill%
  \caption{\label{fig:hardware_comms_denied}The number of visited cells as a function of time illustrates the advantage of the
    decentralized GUTS approach compared to the naive coverage
    planner. Results obtained from flight data at the test site in
    Bloomingdale, OH. The flyable area for this experiment is
    \SI{72100}{\meter\squared}.
  }
  \ifthenelse{\equal{\arxivmode}{true}}{
  \subfloat[Area visited over time\label{sfig:hardware_plot_comms_enabled}]{\includegraphics{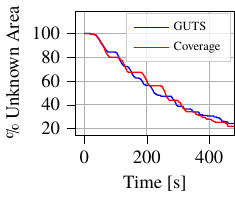}}\hfill%
  }{
  \subfloat[Area visited over time\label{sfig:hardware_plot_comms_enabled}]{\input{figures/hardware/renegade_ridge_231212-comms_enabled_two_robots_percentage.tex}}\hfill%
  }
  \subfloat[Coverage trajectories\label{sfig:hardware_coverage_comms_enabled}]{\includegraphics[width=0.3\linewidth,trim=0 40 20 55,clip]{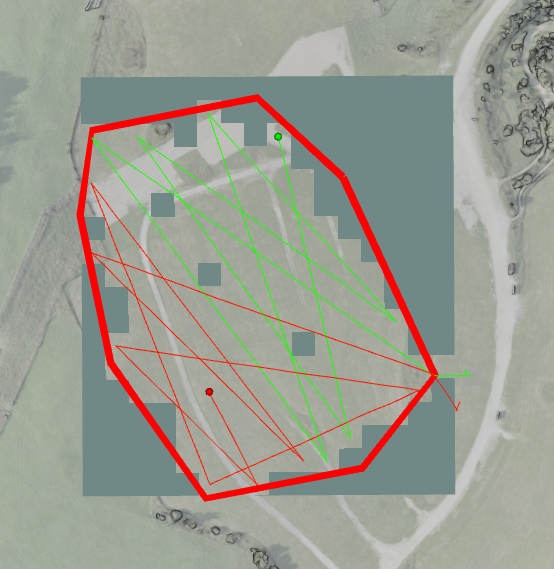}}\hfill%
  \subfloat[GUTS trajectories\label{sfig:hardware_nats_comms_enabled}]{\includegraphics[width=0.3\linewidth,trim=20 50 0 40,clip]{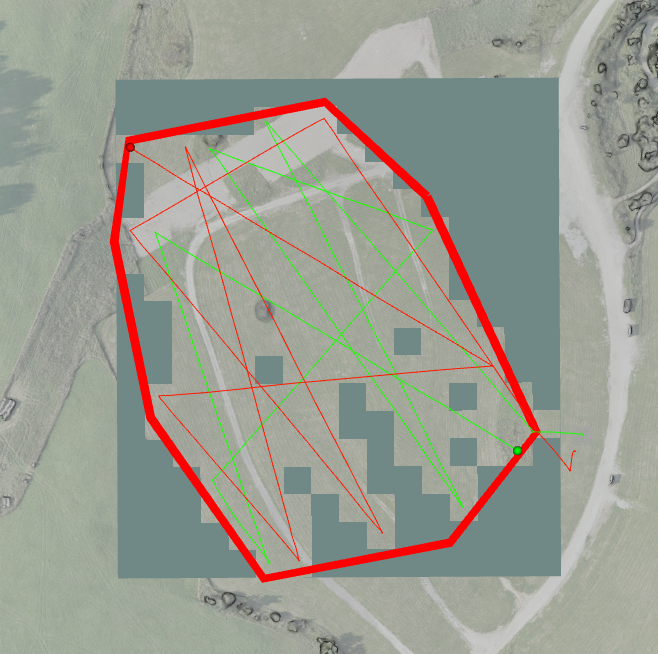}}\hfill%
  \caption{The effect of enabling communication between two robots
    is measured in this hardware experiment.
    The trajectories are shown in red and green.
    The explored area is
    shown as transparent cells, which enables the viewer to see the
    surface terrain. Unknown cells are gray-green and do not enable
    the viewer to see to the terrain below. The GUTS planner performs
    competitively with the greedy coverage planner. The flyable area for
    this experiment is \SI{40700}{\meter\squared}.}
  \subfloat[\label{sfig:tracks_trajs}]{\includegraphics[height=3.8cm,trim=0 0 0 0,clip]{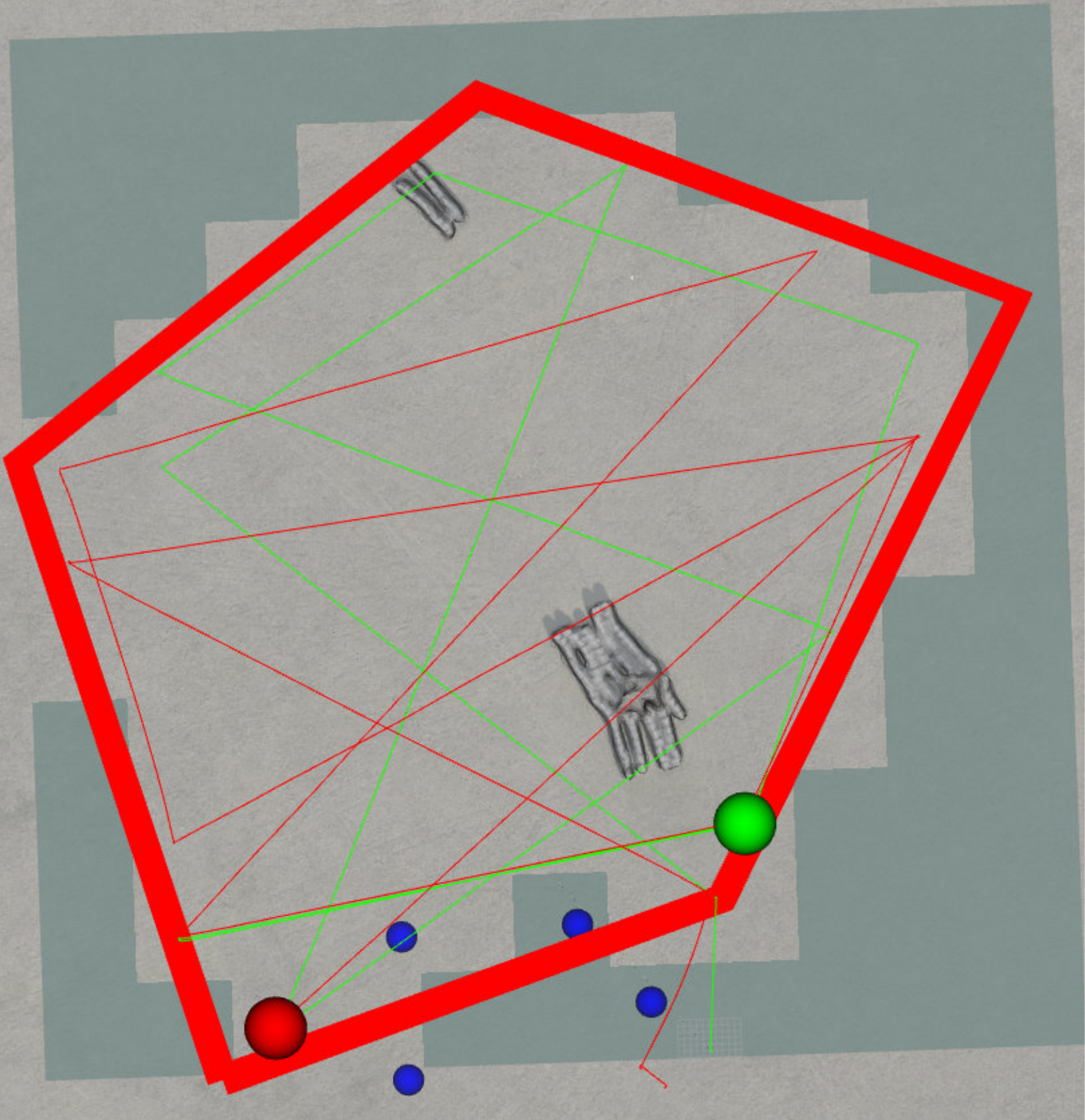}}\hfill%
  \subfloat[\label{sfig:tracks_labeled}]{\includegraphics[height=3.8cm,trim=0 0 0 0,clip]{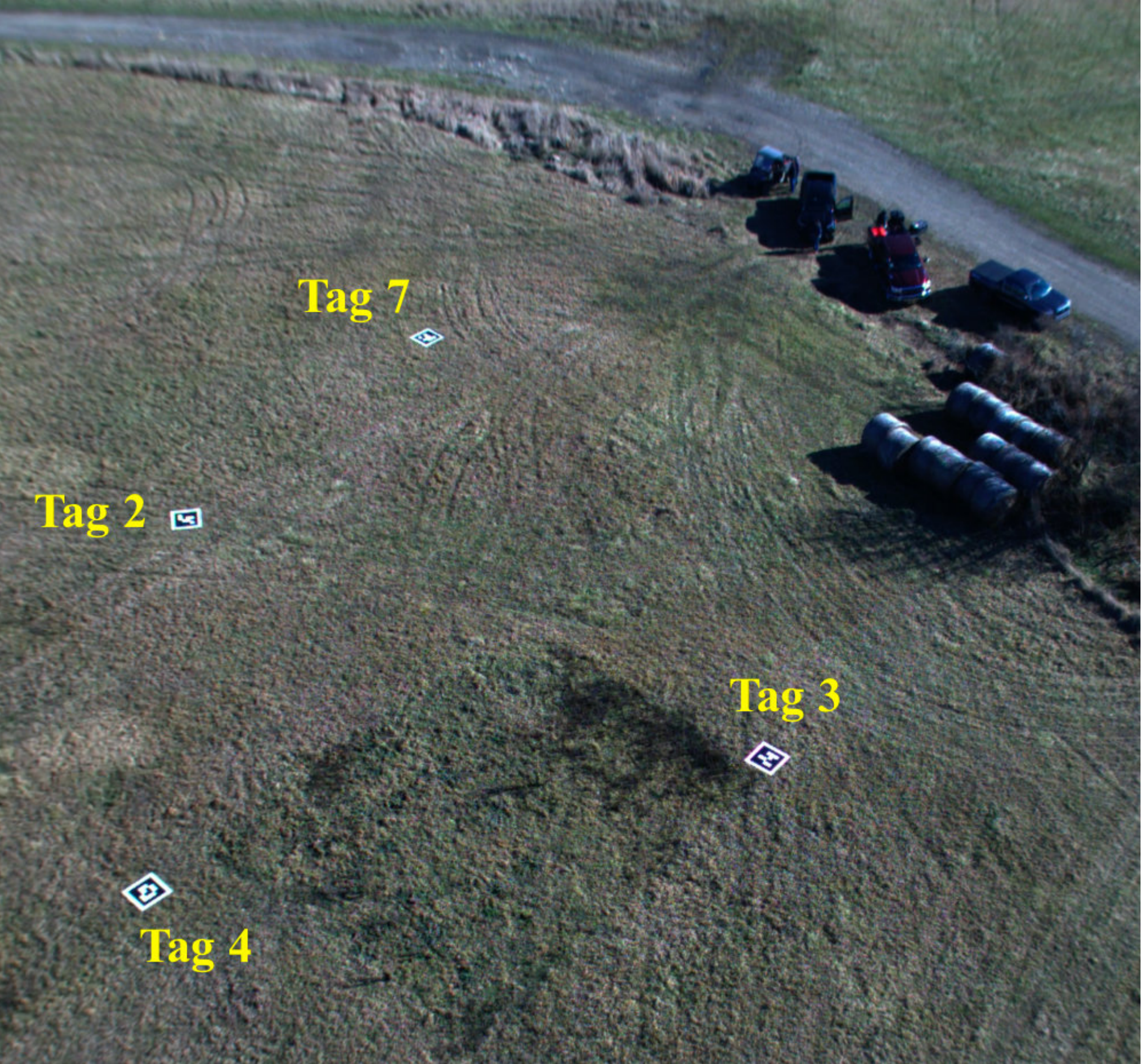}}\hfill%
  \subfloat[Tag 2\label{fig:tag2}]{\includegraphics[width=3.8cm]{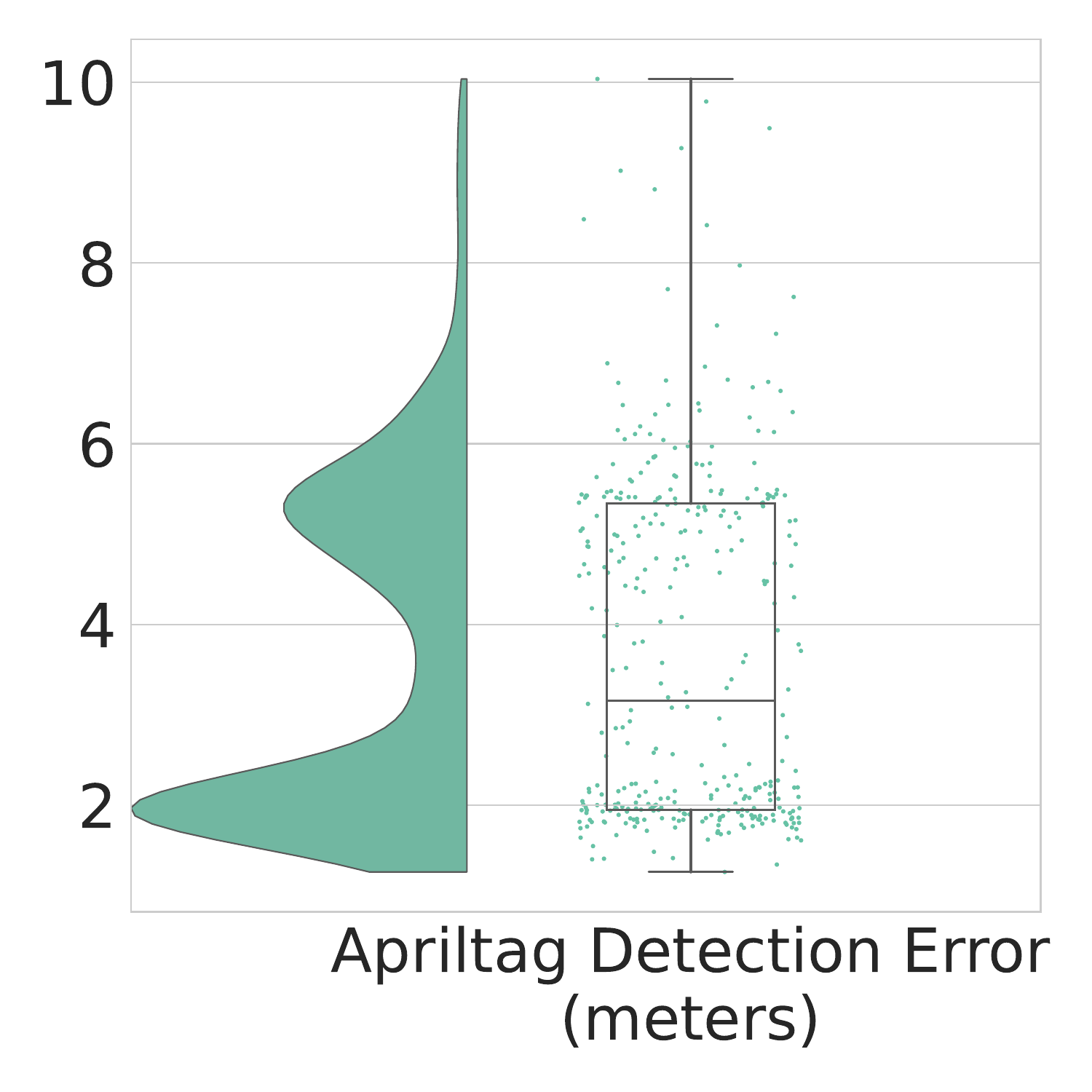}}\hfill%
  \caption{\label{fig:hardware_tracks} For this hardware experiment,
    apriltags are added in the zone and communications are enabled between
    robots.  \protect\subref{sfig:tracks_labeled} provides a
    visualization of the tag positions. Tags 2 and 4 are within the zone
    outlined in red in~\protect\subref{sfig:tracks_trajs}. The accuracy
    with which tag 2 is detected is shown in~\protect\subref{fig:tag2}. The flyable
    area for this experiment is \SI{15000}{\meter\squared}.}
\end{figure}
\begin{figure}
  \begin{minipage}[t]{0.48\textwidth}
  \subfloat[\label{sfig:zone_march_demo}]{\includegraphics[height=7cm,trim=210 60 190 70,clip]{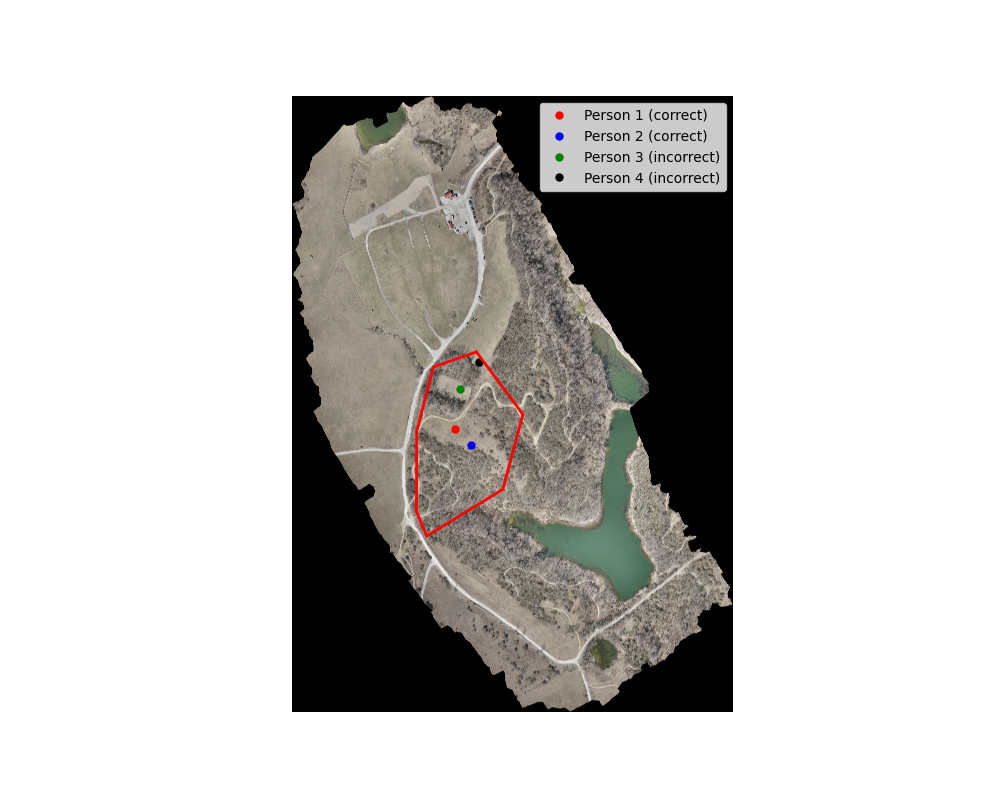}}
  \end{minipage}\hfill
  \begin{minipage}[t]{0.48\textwidth}
    \centering
    \subfloat[Person 1\label{sfig:person1}]{\includegraphics[width=0.48\linewidth]{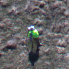}}\hfill%
    \subfloat[Person 2\label{sfig:person2}]{\includegraphics[width=0.48\linewidth]{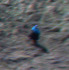}}\\
    \subfloat[Person 3\label{sfig:person3}]{\includegraphics[width=0.48\linewidth]{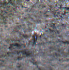}}\hfill%
    \subfloat[Person 4\label{sfig:person4}]{\includegraphics[width=0.48\linewidth]{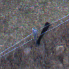}}%
  \end{minipage}\hfill
  \caption{\label{fig:persons}
    Four persons were detected
      within the~\protect\subref{sfig:zone_march_demo} search
      zone for this hardware experiment.~\protect\subref{sfig:person1} and~\protect\subref{sfig:person2}
      are the two correctly detected mannequins hidden in the
      environment.~\protect\subref{sfig:person3}
      and~\protect\subref{sfig:person4} are incorrect detections. A video of
    this experiment may be found at \url{https://youtu.be/qhJS2JhdbAE}.}
  \vspace{-0.5cm}
\end{figure}

The first hardware experiment evaluates the performance of the GUTS
and coverage planners when communication is disabled with a team of
two robots.~\Cref{fig:hardware_comms_denied} provides quantitative
analysis on hardware that validates the performance of the simulation.
In these results, we see that that the GUTS
planner outperforms the coverage planner when communication is
denied. The advantage stems from the stochasticity of the GUTS
planner.  The areas visited by both robots are shown as transparent
cells
in~\cref{sfig:hardware_comms_denied_coverage_map,sfig:hardware_comms_denied_nats_map}.

The second hardware experiment evaluates the performance of the GUTS
and coverage planners when communication is enabled.  A team of two
robots searches the areas shown
in~\cref{sfig:hardware_coverage_comms_enabled,sfig:hardware_nats_comms_enabled}.
They share information about poses and goals; however, there are no
objects to detect in the environment so no information about tracks is
shared.~\Cref{sfig:hardware_plot_comms_enabled} illustrates the
coverage as a function of time. We see that when communications
between robots are enabled, the coverage approach performs similarly
to the GUTS approach, which is in line with what we expect
from the simulation results.

The third field test integrates targets and is shown
in~\cref{fig:hardware_tracks}. Communication is enabled for this
trial. These results correspond to the environment shown
in~\cref{sfig:wingtratopdown}. The targets are shown as blue dots
in~\cref{sfig:tracks_trajs} and labeled with their IDs
in~\cref{sfig:tracks_labeled}. Tags 2 and 4 are contained within the
boundary of the red zone.  Tags 3 and 7 are detected on landing, but
are outside the boundary of the red zone and are not included as a
result. The raw localization errors (i.e., no CPF) for a representative tag is shown
in~\cref{fig:tag2}.  To localize the targets in the global
coordinate frame, a ray from the camera is projected to the apriltag
and intersected with a plane, which represents the environment.
For this test a single trial is conducted with the GUTS
planner. The track certainty is fixed to be low
($c=0.005$). Apriltags are used as targets because they
are detected more reliably compared to neural-based object detectors.

The final field test evaluates the effect of incorporating
cross-platform fusion (CPF) from~\cref{sec:cpf} using static
mannequins as substitutes for persons in the environment. For
all prior experiments, the first sensing payload (Xavier payload) is
used, but for this experiment the second sensing payload (Orin NX
payload) is used. For this experiment, we use two aerial system to
simultaneously cover the search zone illustrated as the red convex
polygon in~\cref{sfig:zone_march_demo}. Two mannequins are hidden in
the environment. No prior knowledge is provided to the system about
the number or type of targets in the environment.  The CPF system
correctly detects both mannequins in the environment
(\cref{sfig:person1,sfig:person2}) and incorrectly detects two
persons (\cref{sfig:person3,sfig:person4}). The localization accuracy
for person 1 is \SI{2.9}{\meter} and \SI{3.6}{\meter} for person
2. The flyable area for this red zone was
\SI{55500}{\meter\squared}. The detector uses a fine-tuned
YOLOv5\footnote{https://github.com/ultralytics/yolov5} model.



\section{Conclusion and Future Work\label{sec:conclusion}}
This paper detailed a system and methodology for decentralized
multi-robot active search and analyzed the performance while varying
the availability of communications and target uncertainty.  A
limitation of the current approach is that it is difficult to scale
with large numbers of robots and environment extents due to the
formulation of the loss function. One way to address this
limitation is to leverage information-theoretic grid compression
strategies that provide a reduced representation while retaining
relevant information. A potential future area of research is to
extend the implementation from 2D to 3D as well as consider terrain
features where objects may be more likely to exist. Robot assisted
search holds the promise of safeguarding lives. To this end, we hope
the results presented in this paper as well as open source software
release accelerate innovation in this area and benefit the robotics
community.


\section{Acknowledgments}
This work was supported in part by the U.S. Army Research Office and
the U.S. Army Futures Command under Contract No. W519TC-23-C-0031.

\section*{References}
{
  \begingroup
  \renewcommand{\bibsection}{}
  \balance
  \scriptsize
  \bibliographystyle{IEEEtranN}
  \bibliography{refs}
  \endgroup
}

\end{document}